\documentclass[journal]{IEEEtran}
\usepackage{graphicx}
\usepackage{amsmath,amssymb} 
\usepackage{color}

\usepackage{epsfig}
\usepackage{listings}
\usepackage{framed}

\usepackage{caption,subfig}
\usepackage{mdwlist}
\usepackage[lined, boxed]{algorithm2e}
\usepackage{multirow}
\usepackage{caption}

\begin{document}

\title{Sparse Dictionary-based Attributes for Action Recognition and Summarization}

\author{Qiang Qiu, Zhuolin Jiang, Rama~Chellappa,~\IEEEmembership{Fellow,~IEEE}
\thanks{Q.~Qiu, Z. Jiang, and R.~Chellappa are with the Center for
Automation Research, UMIACS, University of Maryland, College Park,
MD 20742 USA {(e-mail: qiu@cs.umd.edu, \{zhuolin, rama\}@umiacs.umd.edu})}.
}

\maketitle

\begin{abstract}
   We present an approach for dictionary learning of action attributes via information maximization.  We unify the class distribution and appearance information into an objective function for learning a sparse dictionary of action attributes. The objective function maximizes the mutual information between what has been learned and what remains to be learned
   in terms of appearance information and class distribution for each dictionary atom.
   We propose a Gaussian Process (GP) model for sparse representation to optimize the dictionary objective function. The sparse coding property allows a kernel with compact support in GP to realize a very efficient dictionary learning process. Hence we can describe an action video by a set of compact and discriminative action attributes.
   More importantly, we can recognize modeled action categories in a sparse feature space, which can be generalized to unseen and unmodeled action categories.
   Experimental results demonstrate the effectiveness of our approach in action recognition and summarization.
\end{abstract}

\section{Introduction}

\begin{figure*}
\centering
\includegraphics[angle=0, height=.55\textwidth, width=1\textwidth]{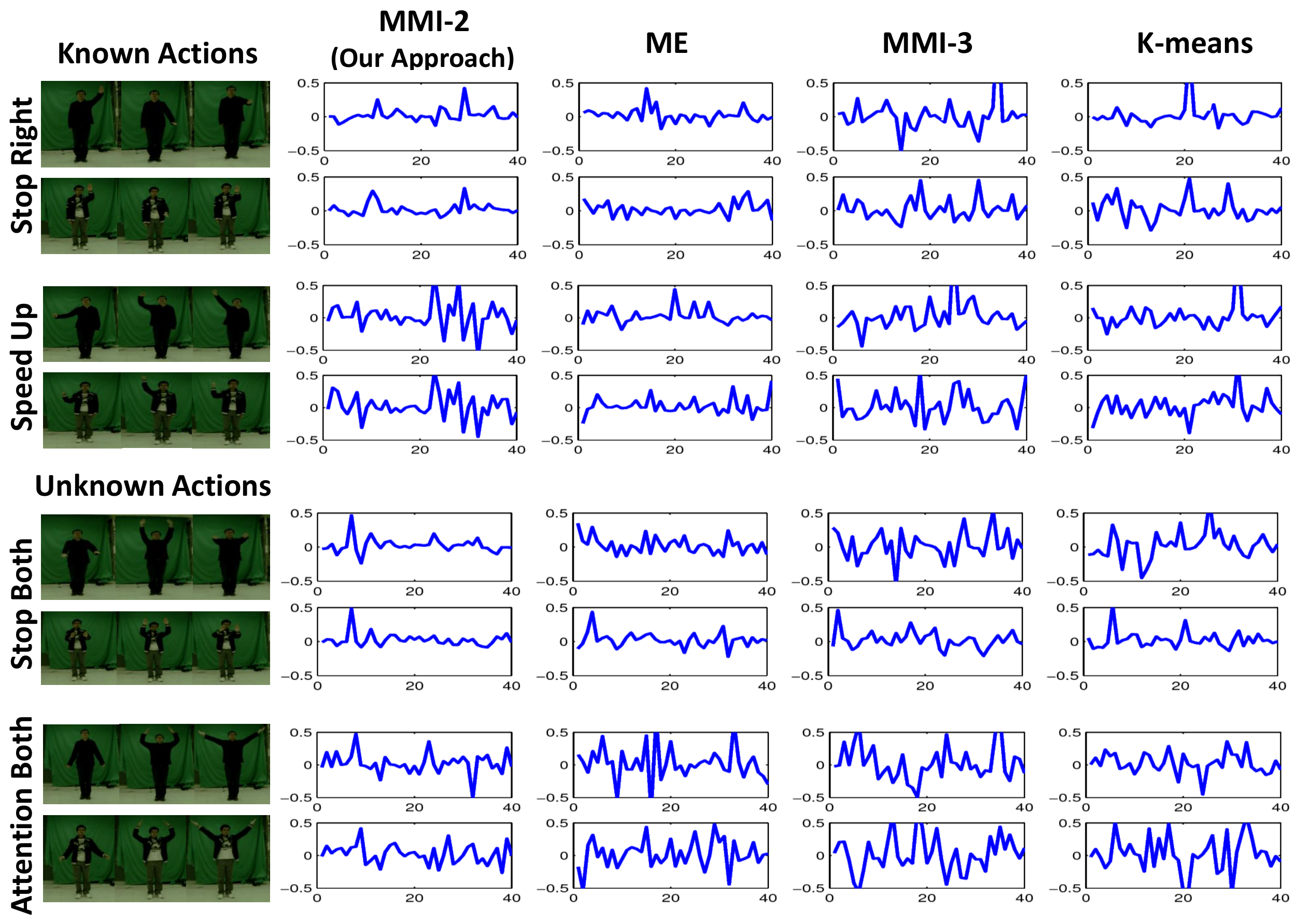}
\caption{
Sparse representations of four actions (two are known and two are unknown to the attribute dictionary) using attribute dictionaries learned by different methods. Each action is performed by two different humans. For visualization purpose, each waveform shows the average of the sparse codes of all frames in an action sequence.
We learned several attribute dictionaries using methods including our approach,  the Maximization of Entropy approach (ME), the MMI-3 approach motivated by \cite{Liu08} and the K-means approach.
A compact and discriminative attribute dictionary should encourage actions from the same class to be described by a similar set of attributes, i.e., similar sparse codes.
The attribute dictionary learned by our approach provides similar waveforms, which shows consistent sparse representations, for the same class action sequences.
 }
 \label{Fig:sparsecodeplot}
\end{figure*}

Describing human actions using attributes is closely related to representing an object using attributes~\cite{Farhadi10}.
Several studies have investigated the attribute-based approaches for object recognition problems~\cite{Lampert09,Farhadi09,Farhadi10,Ramirez10,Yu10}. These methods have demonstrated that attribute-based approaches can not only recognize object categories, but can also describe unknown object categories. In this paper, we propose a dictionary-based approach for learning human action attributes which are useful to model and recognize known action categories, and also describe unknown action categories.

Dictionary learning is one of the approaches for learning attributes (i.e., dictionary atoms) from a set of training samples. In~\cite{Elad_KSVD}, a promising dictionary learning algorithm, K-SVD, is introduced to learn an over-complete dictionary. Input signals can then be represented as a sparse linear combination of dictionary atoms. K-SVD only focuses on focus on representational capability, i.e., minimizes the reconstruction error. The method of optimal direction (MOD)~\cite{Engan99} shares the same sparse coding as K-SVD. ~\cite{Wright09} manually selects training samples to construct a dictionary. ~\cite{Mairal08a} trains one dictionary for each class to obtain discriminability.

Discriminative dictionary learning is gaining attention in many disciplines.
Discriminative K-SVD in~\cite{Zhang10} extends K-SVD by incorporating the classification error into the objective function to obtain a more discriminative dictionary. ~\cite{Pham08} aims to obtain the discriminative power of dictionary by iteratively updating the dictionary from the results of a linear classifier.
\cite{lcksvd} introduces a label consistent constraint to obtain the  discrimination of sparse codes among the classes.
Some other examples include LDA-based basis selection~\cite{ramadic},
distance matrix learning~\cite{Bilenko04}, hierarchical pairwise merging of visual words~\cite{Wang08}, maximization of mutual information (MMI)~\cite{Lazebnik09, Slonim00, Liu08}, and sparse coding-based dictionary learning~\cite{ Mairal08a, superviseddic}.

Recent dictionary-based approaches for learning action attributes include
agglomerative clustering~\cite{Thurau08}, forward selection~\cite{Weinland08} and probabilistic graphical model~\cite{Elgammal03}.~\cite{Liyi10} proposes an unsupervised approach and uses $l_1$ minimization to find basic primitives to represent human motions.

 In this paper, we propose an approach for dictionary learning of human action attributes via information maximization. In addition to using the appearance information between dictionary atoms, we also exploit class label information associated with dictionary atoms to learn a compact and discriminative dictionary for human action attributes. The mutual information for appearance information and class distributions between the learned dictionary and the rest of the dictionary space are used to define the objective function,
  which is optimized using a Gaussian Process (GP) model~\cite{Rasmussen06} proposed for sparse representation.
 The property of sparse coding naturally leads to a kernel with compact support, i.e., zero values for a most portion, in GP for significant speed-ups.  Representation and recognition of actions are accomplished through sparse coefficients related to learned attributes.

 Unlike previous dictionary learning methods that mostly consider learning reconstructive dictionaries, our algorithm can encourage dictionary compactness and discriminability simultaneously. Sparse representation over a dictionary with coherent atoms has the multiple representation problem \cite{shapiro_ima}.
A compact dictionary consists of incoherent atoms, and encourages similar signals, which are more likely from the same class, to be consistently described by a similar set of atoms with similar coefficients.
A discriminative dictionary encourages signals from different classes to be described by either a different set of atoms, or the same set of atoms but with different coefficients \cite{shapiro_ima,sparse_nips, Mairal08a}.
Both aspects are critical for action classification using sparse representation.
As shown in Fig.~\ref{Fig:sparsecodeplot}, our approach produces consistent sparse representations for the same class of signals.

 Our approach adopts the rule of Maximization of Mutual Information to obtain a compact and discriminative dictionary. The dictionary atoms are considered as attributes in our paper. Compared to previous methods, our approach maximizes the mutual information for both the appearance information and class distribution of dictionary atoms to learn a dictionary while~\cite{Slonim00} and~\cite{Liu08} only maximize the mutual information for class distribution. Thus, we can expect improved dictionary compactness from our approach.
Both~\cite{Slonim00} and~\cite{Liu08} obtain a dictionary through merging of two visual words, which can be time-consuming when the dictionary size is large.
Besides, our approach is efficient because the dictionary is learned in the sparse feature space so we can leverage the property of sparse coding to use kernel locality for speeding up the dictionary learning process.

 Our main contributions are:
  \begin{itemize*}
 \item We propose a novel probabilistic model for sparse representation.
 \item We learn a compact and discriminative dictionary for sparse coding via information maximization.
 \item  We describe and recognize human actions, including unknown actions, via a set of human action attributes in a sparse feature space.
 \item We present a simple yet near-optimal action summarization method.
 \end{itemize*}

The rest of this paper is structured as follows. In Sec.~\ref{sec:feature}, we discuss human action features and attributes. We then propose a novel probabilistic model for sparse representation in Sec.~\ref{sec:probmodel}.
In Sec.~\ref{sec:learning}, we present our attribution dictionary learning framework.
We describe how to adopt our attribution dictionary learning method for action summarization in Sec.~\ref{sec:sampling}.
Experimental results are given in Sec.~\ref{sec:experiment} to demonstrate the effectiveness of our approach for action recognition and summarization.

\section{Action Features and Attributes}
\label{sec:feature}
Human action features are extracted from an action interest region for representing and describing actions. The action interest region is defined as a bounded region around the human performing the activity, which is obtained using background subtraction and/or tracking.

\subsection{Basic Features}

The human action attributes require feature descriptors to represent visual aspects.  We introduce basic features, including both local and global features, used in the paper.

\textbf{Global Features:} Global features encode rich information from an action interest region, so they generally perform better than local features in recognition.
When cameras and backgrounds are static, we use the silhouette-based feature descriptor presented in~\cite{Lin09} to capture shape information,
while we use Histogram of oriented gradient (HOG) descriptors used in~\cite{Dalal05} for dynamic backgrounds and moving cameras. For encoding motion information, we use optical-flow based feature descriptors as in~\cite{Efros03}. We use Action Bank descriptors introduced in~\cite{actbank} to demonstrate that our attribute learning method can enhance the discriminability of  high-level global features.

\textbf{Local Features:} Spatio-temporal local features describe a video as a collection of independent patches or $3$D cuboids, which are less sensitive to viewpoint changes, noise and partial occlusion. We first extract a collection of space-time interest points (STIP) introduced in~\cite{Laptev08} to represent an action sequence, and then use HOG and histogram of flow to describe them.

\subsection{Human Action Attributes}
Motivated by~\cite{Thurau08,Weinland08,Elgammal03}, an action can be represented as a set of basic action units. We refer to these basic action units as human action attributes. In order to effectively describe human actions, we need to learn a representative and semantic set of action attributes. Given all the basic features from training data, we aim to learn a compact and discriminative dictionary where all the dictionary atoms can be used as human action attributes. The final learned dictionary can be used as a ``Thesaurus" of human action attributes.
Each human action is then decomposed as sparse linear combinations of attributes in the thesaurus though sparse coding.
The sparse coefficient associated with each attribute measures its weight in representing an action.

\section{A Probabilistic Model for Sparse Representation}
\label{sec:probmodel}

Before we present our dictionary learning framework, we first suggest a novel probabilistic model for sparse representation motivated by~\cite{Guestrin08}.

\subsection{Reconstructive Dictionary Learning}

A reconstructive dictionary can be learned through K-SVD~\cite{Elad_KSVD}, which is a method to learn an over-complete dictionary for sparse coding. Let $Y$ be a set of $N$ input signals in a $n$-dimensional feature space  $Y=[y_1...y_N], ~y_i \in \mathbb{R}^{n}$.
In K-SVD, a dictionary with a fixed number of $K$ atoms is learned by finding a solution iteratively to the following problem:
\begin{eqnarray}
\arg \min_{D,X}\|Y-DX\|^2_2 & ~~~s.t.  \forall i, \|x_i\|_0 \leq T
\label{eqt:representationerr}
\end{eqnarray}
where $D=[d_1...d_K], ~d_i \in \mathbb{R}^{n}$ ($K>n$) is the learned dictionary, $X=[x_1,...,x_N], ~x_i \in \mathbb{R}^{K}$ are the sparse codes of input signals $Y$, and $T$ specifies the sparsity that each signal has fewer than $T$ atoms in its decomposition.
Each dictionary atom $d_i$ is $L_2$-normalized.
The learned dictionary $D$ from (\ref{eqt:representationerr}) only minimizes the reconstruction error, so it is not optimal in terms of compactness and discriminability.

\subsection{A Gaussian Process}
\label{sec:gp}

  Given a set of input signals $Y$, $Y=[y_1...y_N], y_i \in \mathbb{R}^{n}$, there exists an infinite dictionary space $\mathcal{D} \subseteq \mathbb{R}^n$. Each dictionary atom $d_i \in \mathcal{D}$ maps the set of input signals to its corresponding sparse coefficients $x_{d_i} = [x_{i,1}...x_{i,N}]$ in $X$, which can be viewed as its observations to the set of input signals.
When two dictionary atoms $d_i$ and $d_j$ are similar, it is more likely that input signals will use them simultaneously in their sparse decomposition ~\cite{Ramirez10}. Thus the similarity of two dictionary atoms can be assessed by the correlation between their observations (i.e., sparse coefficients). Such correlation property of sparse coefficients has been used in ~\cite{Ramirez10} to cluster dictionary atoms.

With the above formulation, we obtain a problem which is commonly referred as a GP model. A GP is specified by a mean function and a symmetric positive-definite covariance function $\mathcal{K}$.
Since we simplify our problem by assuming an initial dictionary $D^o$, we only need to specify entries in the covariance function $\mathcal{K}$ for atoms existing in $D^o$, and leave the rest undefined.
In this paper, for each pair of dictionary atoms $\forall d_i, d_j \in D^o$, the corresponding covariance function entry $\mathcal{K}(i, j)$ is defined as the covariance between their associated sparse coefficients $cov(x_{d_i}, x_{d_j})$.
For simplicity, we use the notation $\mathcal{K}_{(d_i, d_j)}$ to refer to the
covariance entry at the indices of $d_i$, $d_j$. Similarly,  we use $\mathcal{K}_{(D^*, D^*)}$ to denote the covariance matrix for a set of dictionary atoms $D^*$.

The GP model for sparse representation provides the following useful property:
given a set of dictionary atoms $D^*$ and the associated sparse coefficients $X_{D^*}$,
the distribution $P(X_{d^*} | X_{D^*} )$ at any given testing dictionary atom $d^*$ is a Gaussian with a closed-form conditional variance~\cite{Rasmussen06}.
\begin{eqnarray}
\mathbb{V}(d^* | D^*) = \mathcal{K}_{(d^*, d^*)} - \mathcal{K}_{(d^*, D^*)}^T \mathcal{K}_{(D^*, D^*)}^{-1} \mathcal{K}_{(d^*, D^*)}
\label{eqt:condvar}
\end{eqnarray}
where $\mathcal{K}_{(d^*, D^*)}$ is the vector of covariances between $d^*$ and each atom in $D^*$.

\subsection{Dictionary Class Distribution}
\label{sec:label-dist}
When the set of input signals $Y$ is labeled with one of $M$ discrete class labels, we can further derive class related distributions over sparse representations.

As mentioned, each dictionary atom $d_i$ maps the set of input signals to its corresponding sparse coefficients $x_{d_i} = [x_{i,1}...x_{i,N}]$ in $X$. Since each coefficient $x_{i,j}$ here corresponds to an input signal $y_j$, it is associated with a class label.
If we aggregate $x_{d_i}$ based on class labels, we obtain a $M$ sized vector. After normalization, we have the conditional probability $P(L |d_i), ~L \in [1,M]$, where  $P(L |d_i)$  represents
  the probability of observing a class given a dictionary atom.

\section{Learning Attribute Dictionary}
\label{sec:learning}
As the optimal dictionary size is rarely known in advance, we first obtain through K-SVD an initial dictionary $D^o$ of a large size $K$.
As discussed, the initial dictionary $D^o$ from (\ref{eqt:representationerr}) only minimizes the reconstruction error, and is not optimal in terms of compactness and discriminability.
Then we learn a compact and discriminative dictionary from the initial dictionary via information maximization.

Given the initial dictionary $D^o$ obtained from (\ref{eqt:representationerr}), we aim to compress it into a dictionary $D^*$ of size $k$,  which encourages the signals from the same class to have very similar sparse representations, as shown in Fig.~\ref{Fig:sparsecodeplot}. In other words, the signals from the same class are described by a similar set of attributes, i.e., dictionary atoms. Therefore, a compact and discriminative dictionary is more desirable.

An intuitive heuristic is to start with $D^*=\emptyset$, and iteratively choose the next best atom $d^*$ from $D^o \backslash D^*$ which provides a maximum increase for the entropy of $D^*$, i.e., $\arg \max_{d^*} H(d^*|D^*)$, until $|D^*|=k$, where $D^o \backslash D^*$ denotes the remaining dictionary atoms after $D^*$ have been removed from the initial dictionary $D^o$. Using the GP model, we can evaluate $H(d^*|D^*)$ as a closed-form Gaussian conditional entropy,
\begin{equation}
H(d^*|D^*)=\frac{1}{2}log(2\pi e \mathbb{V}(d^* | D^*))
\label{eqt:varianceentropy}
\end{equation}
where $\mathbb{V}(d^* | D^*)$ is defined in (\ref{eqt:condvar}).
This heuristic is a good approximation to the \emph{maximization of joint entropy} (ME) criteria, i.e.,  $\arg \max_{D^*} H(D^*)$.

With the ME rule, as atoms in the learned dictionary are less correlated to each other due to their high joint entropy, the learned dictionary is compact. However, the maximal entropy criteria will favor attributes associated with the beginning and the end of an action, as they are least correlated.
Such a phenomenon is shown in Fig.~\ref{fig:dicme} and Fig.~\ref{fig:compositionplot} in the experiment section. Thus we expect high reconstruction error and weak discriminability. To mitigate this in our dictionary learning framework, we adopt Maximization of Mutual Information (MMI) as the criteria for ensuring dictionary compactness and discriminability.

\subsection{MMI for Unsupervised Learning (\textbf{MMI-1})}

The rule of maximization of entropy only considers the entropy of dictionary atoms. Instead we choose to learn $D^*$ that most reduces the entropy about the rest of dictionary atoms
$D^o \backslash D^*$.
\begin{eqnarray}
\arg \max_{D^*} I(D^*; D^o \backslash D^*)
 \label{eqt:maxmutualinfo1}
\end{eqnarray}

It is known that maximizing the above criteria is NP-complete. A similar problem has been studied in the machine learning literature~\cite{Guestrin08}.
We can use a very simple greedy algorithm here. We start with $D^*=\emptyset$, and iteratively choose the next best dictionary atom $d^*$ from $D^o \backslash D^*$ which provides a maximum increase in mutual information, i.e.,
\begin{eqnarray}
\arg \max_{d^* \in D^o \backslash D^*} & I(D^* \cup d^*; D^o \backslash (D^* \cup d^*)) - I(D^*; D^o \backslash D^*) \nonumber \\
& = H(d^* | D^*) - H(d^* | \bar{D}^*);
 \label{eqt:MMI1-algo}
\end{eqnarray}
where $\bar{D}^*$ denotes $D^o \backslash (D^* \cup d^*)$.
Intuitively, the ME criteria only considers $H(d^* | D^*)$, i.e., forces $d^*$ to be most different from already selected dictionary atoms $D^*$,  now we also consider $- H(d^* | \bar{D}^*)$ to force $d^*$ to be most representative among the remaining atoms.

It has been proved in~\cite{Guestrin08} that the above greedy algorithm is submodular and serves a polynomial-time approximation that is within $(1-1/e)$ of the optimum. Using arguments similar to the ones presented in~\cite{Guestrin08}, the near-optimality of our approach can be guaranteed if the initial dictionary size $|D^o|$ is sufficiently larger than $2|D^*|$.

Using the proposed GP model, the objective function in (\ref{eqt:MMI1-algo}) can be written in a closed form using (\ref{eqt:condvar}) and (\ref{eqt:varianceentropy}).

\begin{eqnarray}
\arg \max_{d^* \in D^o \backslash D^*} \frac{\mathcal{K}_{(d^*, d^*)} - \mathcal{K}_{(d^*, D^*)}^T \mathcal{K}_{(D^*, D^*)}^{-1} \mathcal{K}_{(d^*, D^*)}}
{\mathcal{K}_{(d^*, d^*)} - \mathcal{K}_{(d^*, \bar{D}^*)}^T \mathcal{K}_{(\bar{D}^*, \bar{D}^*)}^{-1} \mathcal{K}_{(d^*, \bar{D}^*)}}
 \label{eqt:MMI1-obj}
\end{eqnarray}

Given the initial dictionary size $|D^o|=K$, each iteration requires $\mathcal{O}(K^4)$ to evaluate (\ref{eqt:MMI1-obj}). Such an algorithm seems to be computationally infeasible for any large initial dictionary size. The nice feature of this approach is that we model the covariance kernel $\mathcal{K}$ over sparse codes $X$, which entitles $\mathcal{K}$ a compact support, i.e., most entries of $\mathcal{K}$ have zero or very tiny values. After we ignore those zero value portion while evaluating (\ref{eqt:MMI1-obj}), the actual computation becomes very efficient.

\subsection{MMI for Supervised Learning  (\textbf{MMI-2}) }
The objective functions in (\ref{eqt:maxmutualinfo1}) and (\ref{eqt:MMI1-algo}) only consider the appearance information of dictionary atoms, hence $D^*$ is not optimized for classification. For example, attributes to distinguish a particular class can possibly be missing in $D^*$.
 So we need to use appearance information and class distribution to construct a dictionary that also causes minimal loss information about labels.

Let $L$ denote the labels of $M$ discrete values,  $L \in [1,M]$. In Sec.~\ref{sec:label-dist}, we discussed how to obtain $P(L |d^*)$, which represents the probability of observing a class given a dictionary atom.
Give a set of dictionary atom $D^*$, we define $P(L |D^*) = \frac{1}{|D^*|} \sum_{d_i \in D^* } P(L |d_i) $. For simplicity, we denote $P(L |d^*)$ as $P(L_{d^*})$, and $P(L |D^*)$ as $P(L_{D^*})$.

To enhance the discriminative power of the learned dictionary, we propose to modify the objection function (\ref{eqt:maxmutualinfo1}) to
\begin{eqnarray}
\arg \max_{D^*} I(D^*; D^o \backslash D^*) + \lambda I(L_{D^*}; L_{D^o \backslash D^*}) \label{eqt:maxmutualinfo2}
\end{eqnarray}
where $\lambda \ge 0$ is the parameter to regularize the emphasis on appearance or label information.
When we write (\ref{eqt:maxmutualinfo2}) in its approximation version as (\ref{eqt:maxmutualinfo2-greedy})
\begin{eqnarray}
\arg \max_{d^* \in D^o \backslash D^*} [H(d^* | D^*) - H(d^* | \bar{D}^*)] \nonumber \\
+ \lambda [H(L_{d^*} | L_{D^*}) - H(L_{d^*} | L_{\bar{D}^*})]
\label{eqt:maxmutualinfo2-greedy}
\end{eqnarray}
where
\[
H(L_{d^*} | L_{D^*})=-\sum_{L \in [1,M]} P(L_{d*})P(L_{D^*})\log P(L_{d^*})
\]
we can easily notice that now we also force the classes associated with $d^*$ to be most different from classes already covered by selected atoms $D^*$; and at the same time, the classes associated with $d^*$ should be most representative among classes covered by the remaining atoms. Thus the learned dictionary is not only compact, but also covers all classes to maintain the discriminability. It is interesting to note that MMI-1 is a special case of MMI-2 with $\lambda=0$.

The parameters $\lambda$ in (\ref{eqt:maxmutualinfo2-greedy}) are data dependent and can be estimated as the ratio between the maximal information gained from an atom to the respective compactness and discrimination measure, i.e.,
\begin{align} \label{eqt:est-lambda}
\lambda &= \frac{\max_{d^* \in D^o}[H(L_{d^*} | L_{D^*}) - H(L_{d^*} | L_{\bar{D}^*})]}{\max_{d^* \in D^o} [H(d^* | D^*) - H(d^* | \bar{D}^*)]}.
\end{align}
For each term in (\ref{eqt:maxmutualinfo2-greedy}), only the first greedily selected atoms are involved in parameter estimation.  This leads to an efficient process in finding the parameters.

\subsection{MMI using dictionary class distribution (\textbf{MMI-3})}

MMI-1 considers the appearance information for dictionary compactness, and MMI-2 uses appearance and class distribution to enforce both dictionary compactness and discriminability. To complete the discussion, MMI-3, which is motivated by \cite{Liu08}, only considers the dictionary class distribution, discussed in Sec.~\ref{sec:label-dist}, for dictionary discriminability.

In MMI-3, we start with an initial dictionary $D^o$ obtained from K-SVD. At each iteration, for each pair of dictionary atoms, $d_1$ and $d_2$, we compute the MI loss if we merge these two into a new dictionary atom $d^*$, and pick the pair which gives the minimum MI loss. We continue the merging process till the desired dictionary size. The MI loss is defined as,
\begin{eqnarray}
\triangle I(d_1, d_2) & = \underset{L \in [1,M], i=1,2 } \sum p(d_i)p(L|d_i)\log p(L|d_i) \nonumber \\ & -p(d_i)p(L|d_i)\log p(L|d^*)
\end{eqnarray}
where \\
\[p(L|d^*) = \frac{p(d_1)}{p(d^*)}p(L|d_1) + \frac{p(d_2)}{p(d^*)}p(L|d_2) \]
\[p(d^*) = p(d_1)+p(d_2)\]

\section{Action Summarization using MMI-1}
\label{sec:sampling}

Summarizing an action video sequence often considers two criteria: \emph{diversity} and \emph{coverage} \cite{video_precis}. The diversity criterion requires the elements in a summary be as different from each other as possible; and the coverage criterion requires a summary to also represent the original video well.

In (\ref{eqt:MMI1-algo}), the first term $H(d^* | D^*)$ forces $d^*$ to be most different from already selected dictionary atoms $D^*$.  The second term $- H(d^* | \bar{D}^*)$ to force $d^*$ to be most representative among the remaining atoms.
By considering an action sequence as a dictionary, and each frame as a dictionary atom, MMI-1 serves a near-optimal video summarization scheme. The first term in (\ref{eqt:MMI1-algo}) measures diversity and the second term in (\ref{eqt:MMI1-algo}) measures coverage.
The only revision required here is to define the kernel of the Gaussian process discussed in Sec.~\ref{sec:gp} as $\mathcal{K}_{(d_i, d_j)}=d_i^Td_j$.

The advantage in adopting MMI-1 as a summarization/sampling scheme can be summarized as follows: first, MMI-1 is a simple greedy algorithm that can be executed very efficiently. Second,
the MMI-1 provides near-optimal sampling/summarization results, which is within $(1-1/e)$ of the optimum. Such near-optimality is achieved through a submodular objective function that enforces diversity and coverage simultaneously.

\section{Experimental Evaluation}
\label{sec:experiment}

\begin{figure*} [t]
\centering
 \subfloat[Purity] {\label{fig:purity} \includegraphics[angle=0, height=0.25\textwidth, width=0.32\textwidth]{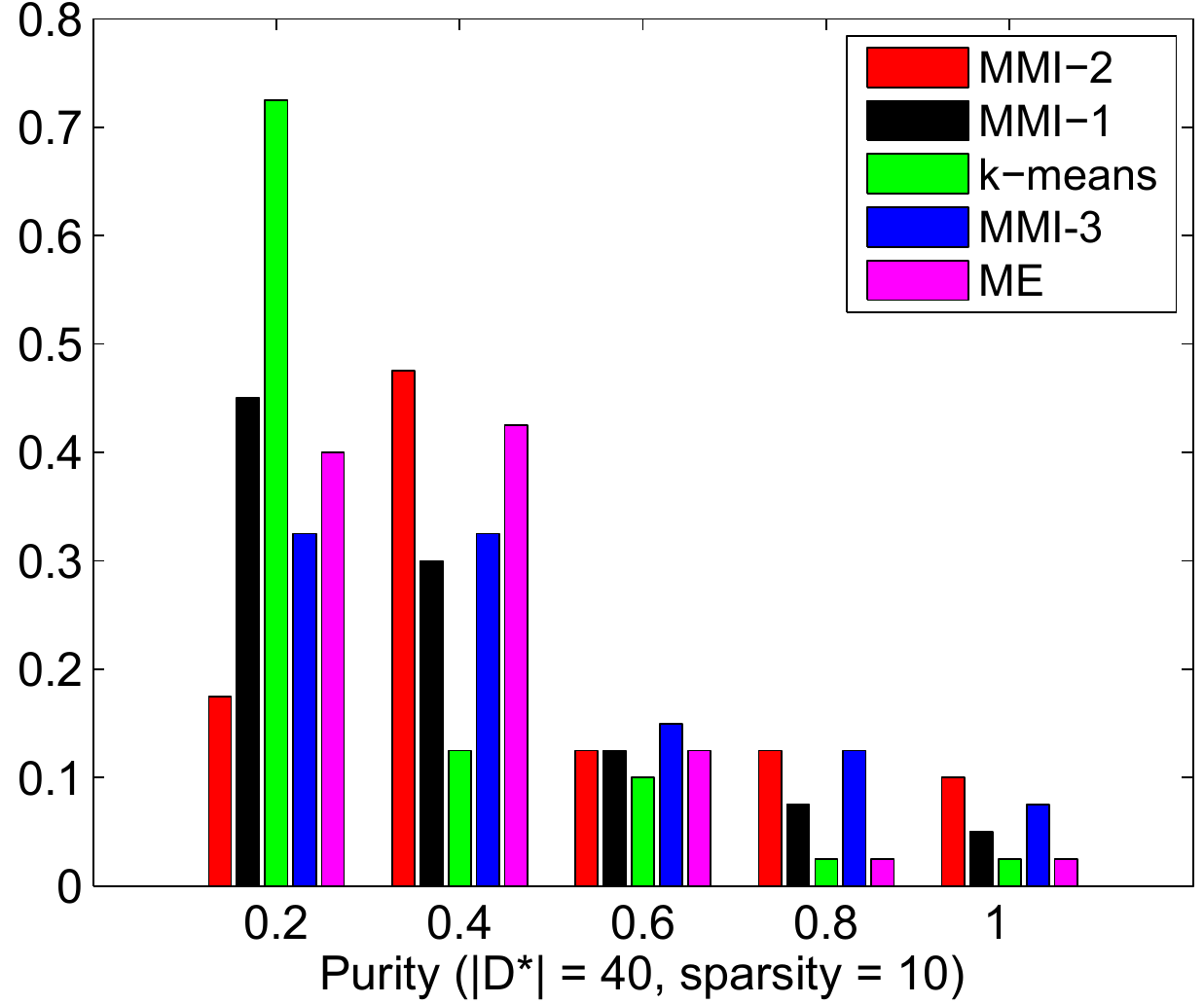} \hspace{36pt}}
  \subfloat[Compactness] {\label{fig:compactness} \includegraphics[angle=0, height=0.25\textwidth, width=0.32\textwidth]{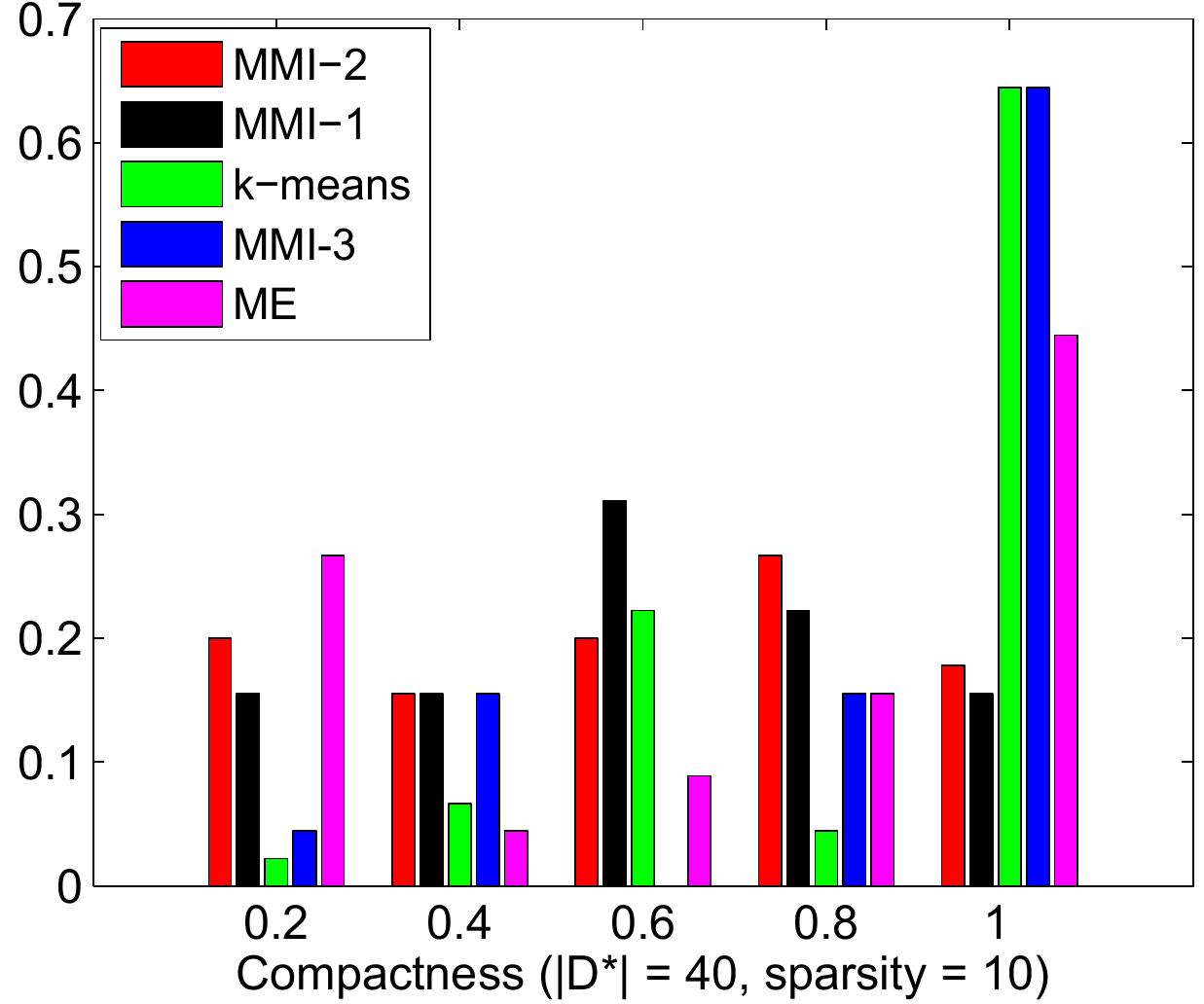}}
\caption{Purity and compactness of learned dictionary $D^*$: purity is the histograms of the maximum probability observing a class given a dictionary atom, and compactness is the histograms of $D^{*T}D^*$. At the right-most bin of the respective figures, a discriminative and compact dictionary should exhibit high purity and small compactness. MMI-2 dictionary is most ``pure" and second most compact (MMI-1 is most compact but much less pure.)}
\label{Fig:purity-compactness}
\end{figure*}

This section presents an experimental evaluation using four public action datasets: Keck gesture dataset~\cite{Lin09}, Weizmann action dataset~\cite{weizmanndata}, UCF sports action dataset~\cite{ucfsportdata}, and UCF50 action dataset~\cite{ucf50}.
On the Keck gesture dataset, we thoroughly evaluate the basic behavior of our proposed dictionary learning approaches MMI-1, MMI-2, and MMI-3, in terms of dictionary compactness and discriminability, by comparing with other alternatives. Then we further evaluate the discriminability of our learned action attributes over the popular Weizmann aciton dataset, the challenging UCF sports and UCF50 action datasets.

\begin{figure*} [ht]
\centering
 \subfloat[MMI-2 shape attributes] {\label{fig:dicno2} \includegraphics[angle=0, height=0.23\textwidth, width=0.27\textwidth]{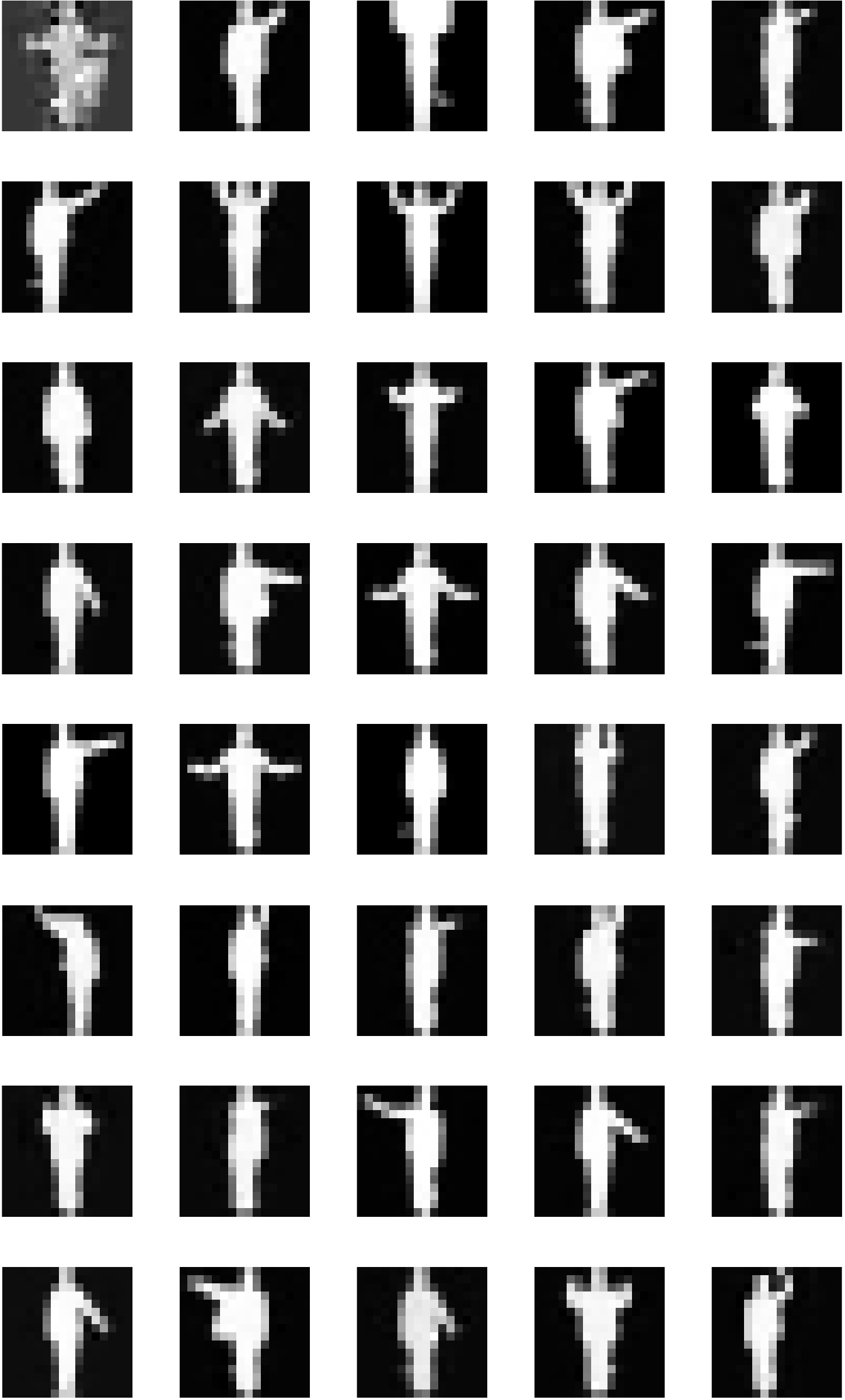} \hspace{13pt}}
  \subfloat[ME shape attributes] {\label{fig:dicme} \includegraphics[angle=0, height=0.23\textwidth, width=0.27\textwidth]{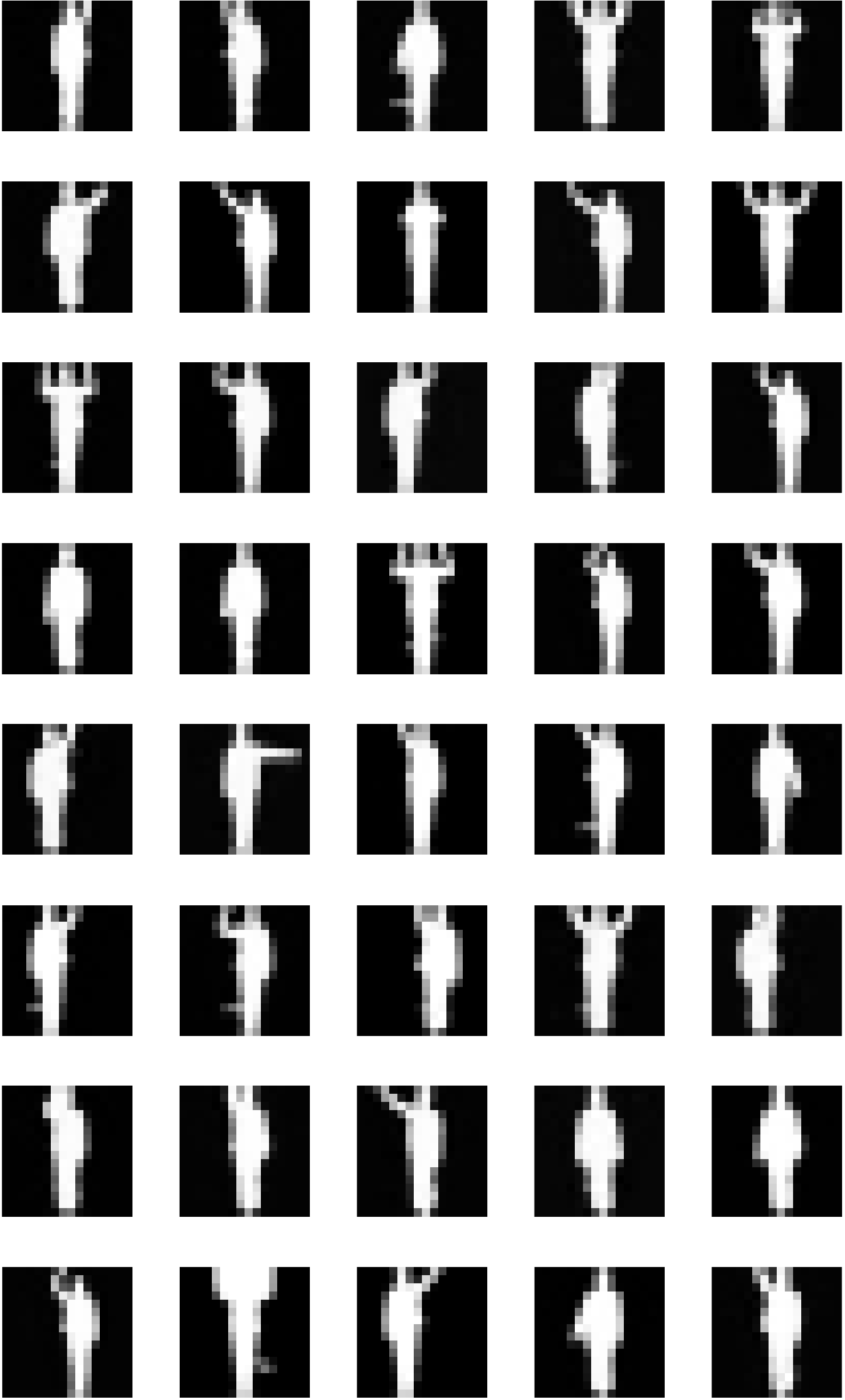} \hspace{13pt}}
  \subfloat[MMI-3 shape attributes] {\label{fig:dicmm} \includegraphics[angle=0, height=0.23\textwidth, width=0.27\textwidth]{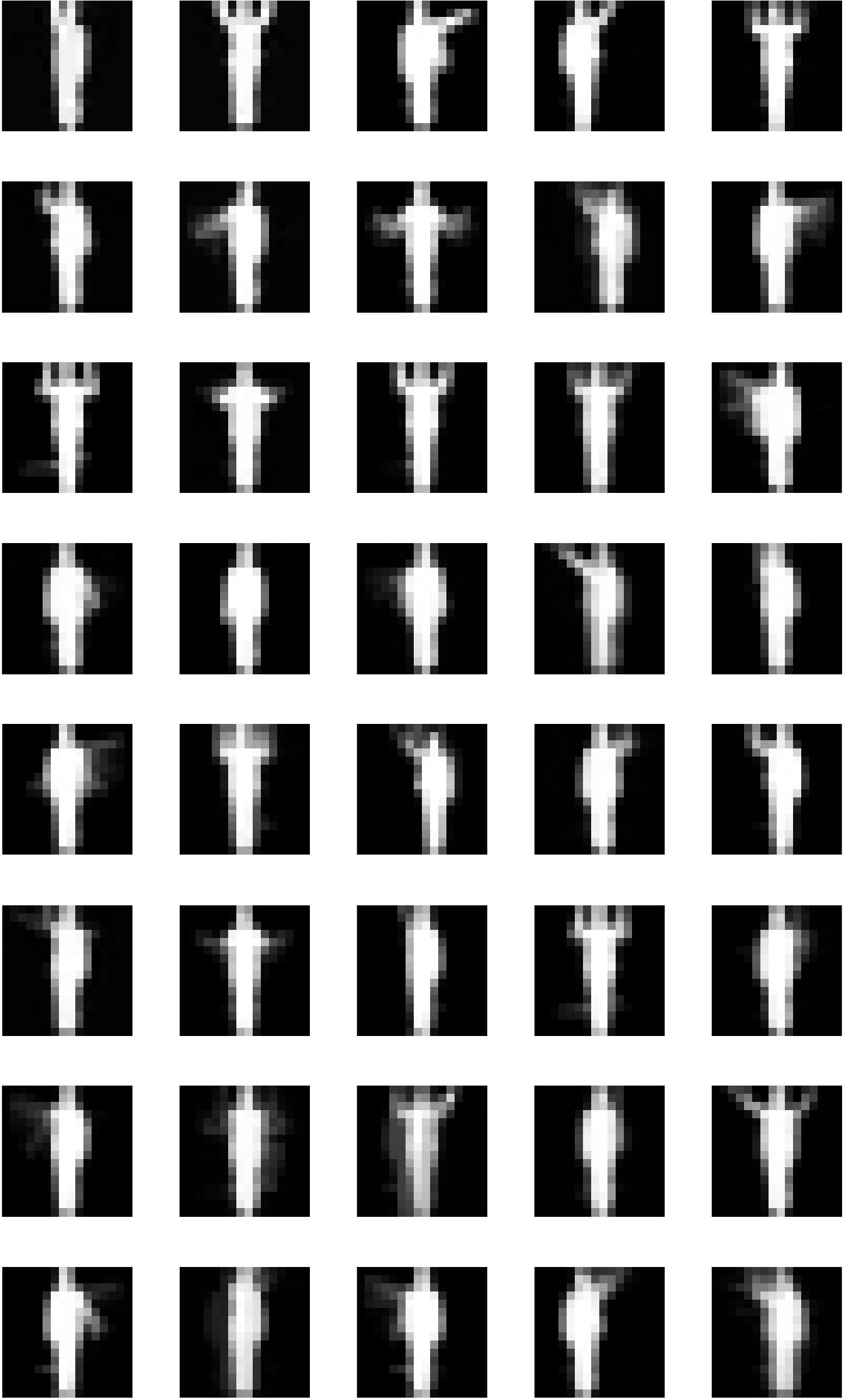}}
\\  \subfloat[Description to two example frames in an unknown action \emph{flap} using attribute dictionaries (Sparsity 10 is used and top-4 attributes are shown.)] {\label{fig:compositionplot} \includegraphics[angle=0, height=0.15\textwidth, width=1\textwidth]{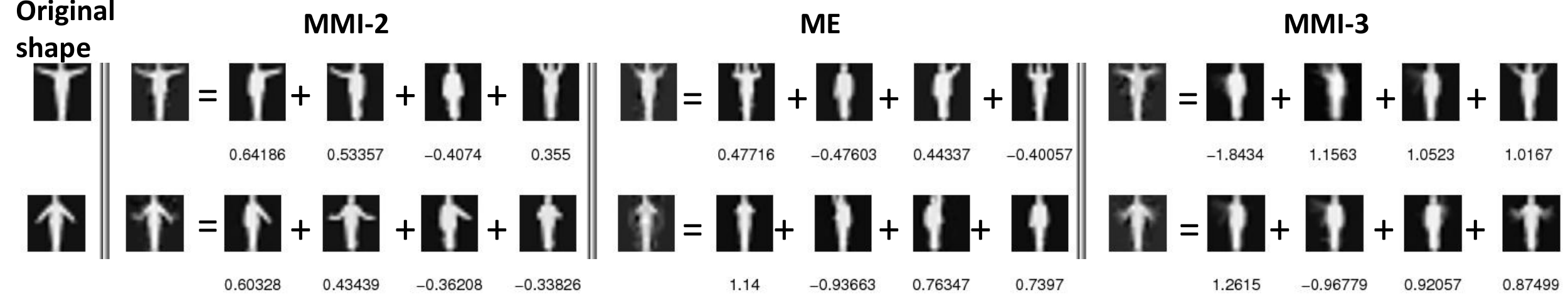}}
\caption{Learned attribute dictionaries on shape features (``unseen" classes: \emph{flap}, \emph{stop both} and \emph{attention both})}
\label{Fig:dic40}
\end{figure*}

\subsection{Comparison with Alternative Approaches}

The Keck gesture dataset consists of 14 different gestures, which are a subset of the military signals. These 14 classes include turn left, turn right, attention left, attention right, flap, stop left, stop right, stop both, attention both, start, go back, close distance, speed up, come near. Each of the 14 gestures is performed by three subjects.  Some sample frames from this dataset are shown in Fig.~\ref{Fig:sparsecodeplot}.

For comparison purposes, in addition to MMI-1, MMI-2 and MMI-3 methods proposed in Sec.~\ref{sec:learning}, we also implemented two additional action attributes learning approaches. The first approach is the maximization of entropy (ME) method discussed before.
The second approach is to simply perform k-means over an initial dictionary $D^o$  from K-SVD to obtain a desired size dictionary.

\begin{figure*} [ht]
\centering
 \subfloat[Shape ($|D^o|=600$)] {\label{fig:shapeacc} \includegraphics[angle=0, height=0.23\textwidth, width=0.25\textwidth]{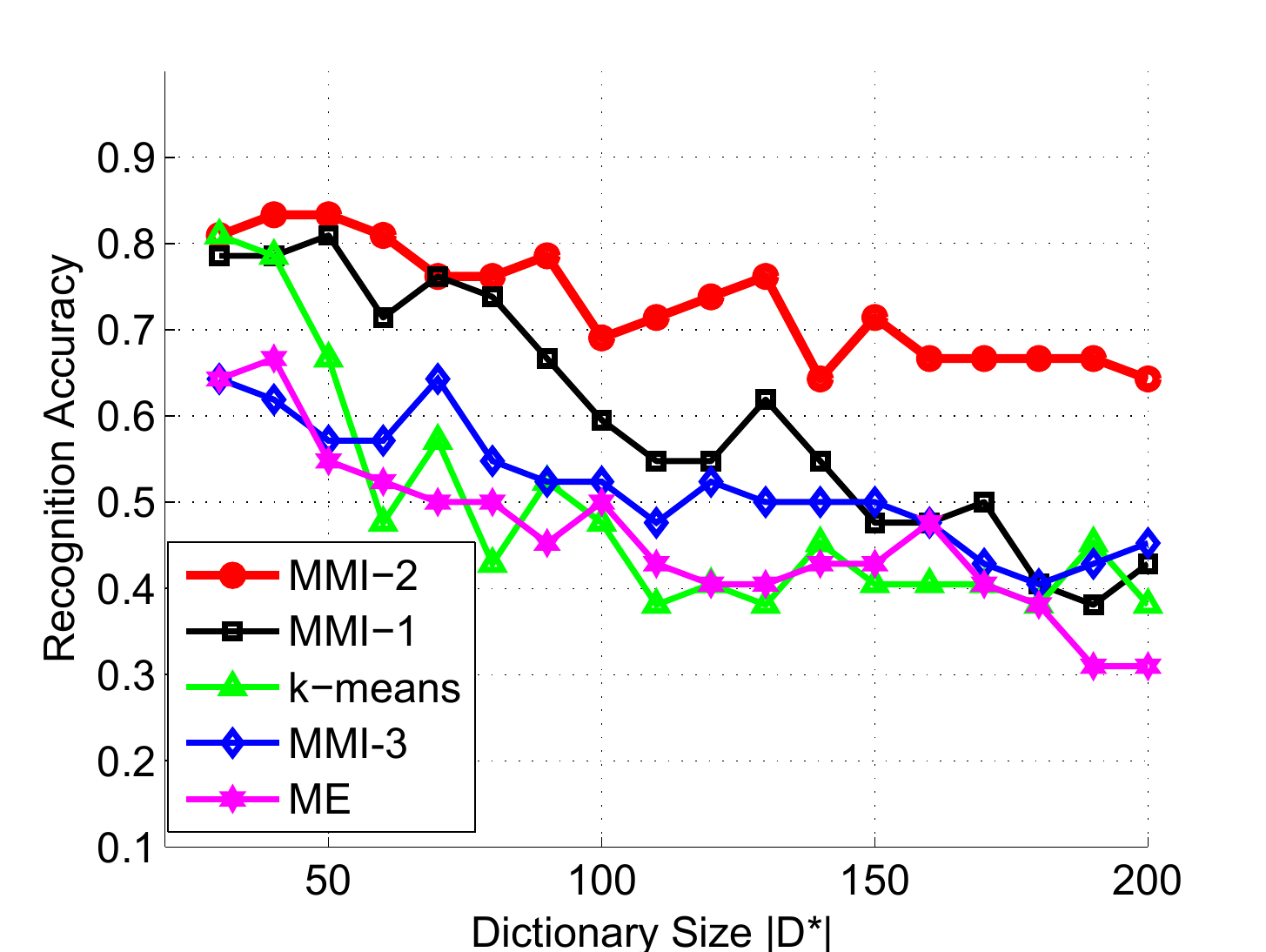}}
  \subfloat[Motion ($|D^o|=600$)] {\label{fig:motionacc} \includegraphics[angle=0, height=0.23\textwidth, width=0.25\textwidth]{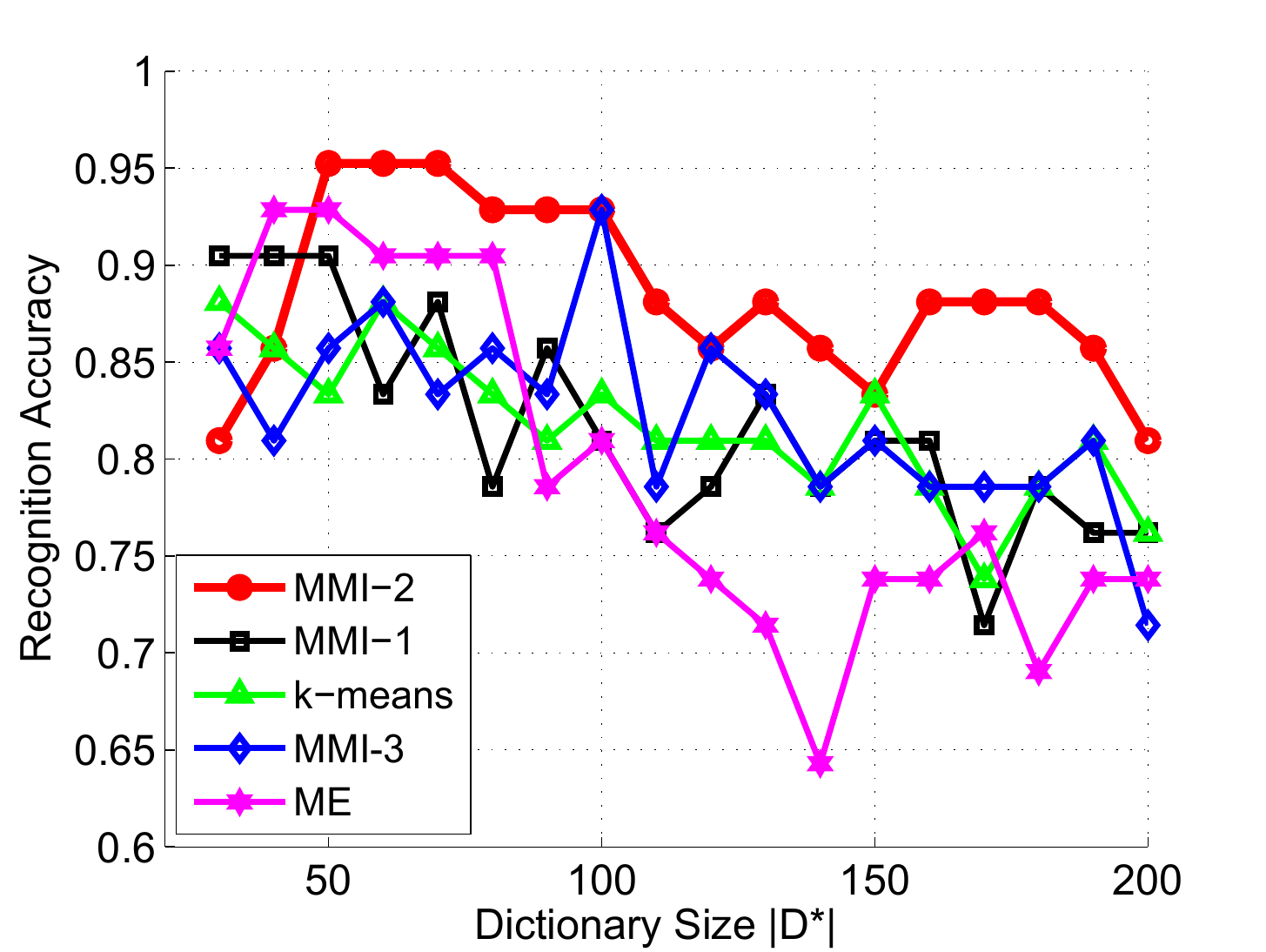}}
  \subfloat[Shape and Motion ($|D^o|=1200$)] {\label{fig:msacc} \includegraphics[angle=0, height=0.23\textwidth, width=0.25\textwidth]{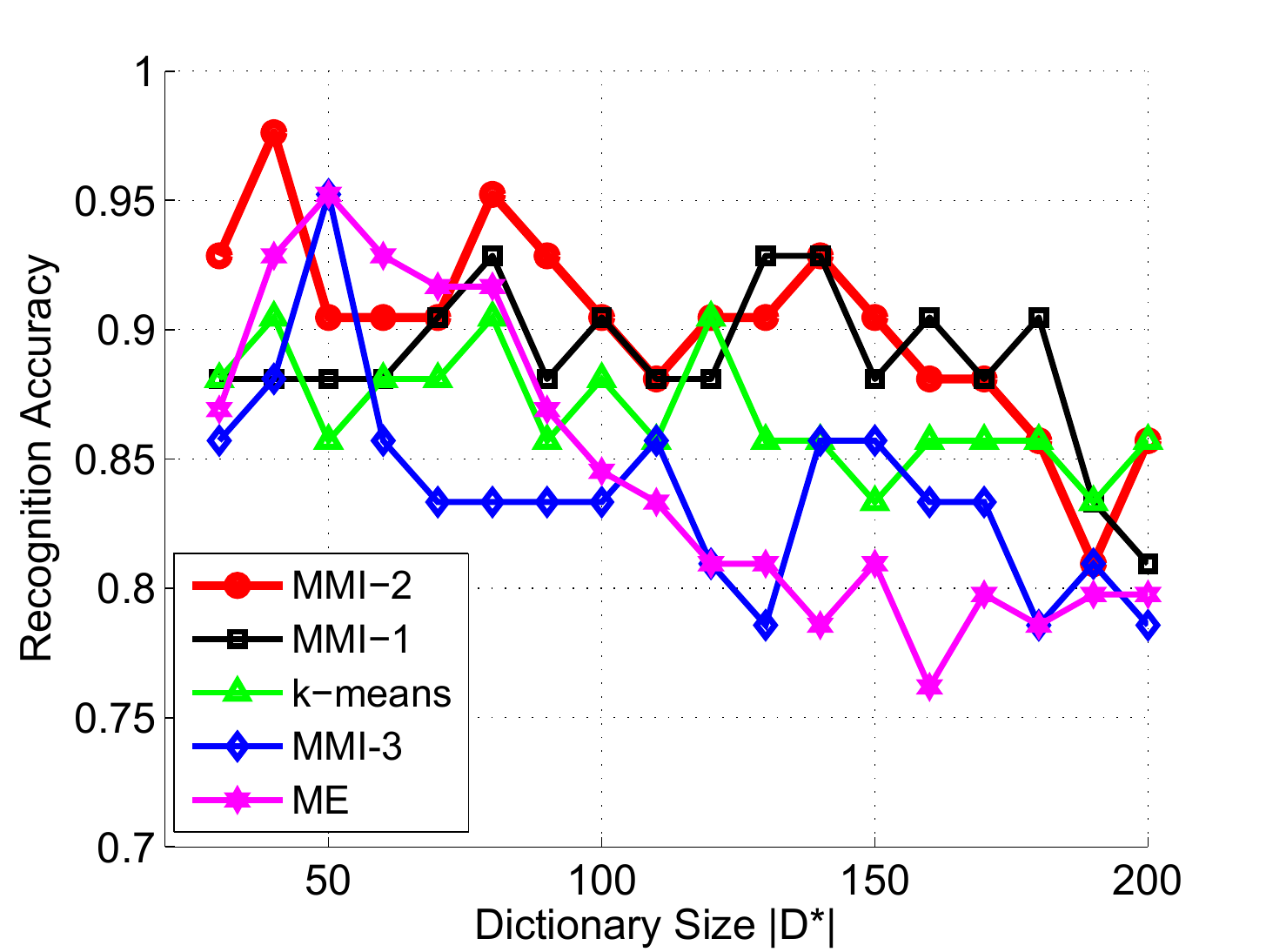}}
  \subfloat[STIP ($|D^o|=600$)] {\label{fig:stipacc} \includegraphics[angle=0, height=0.23\textwidth, width=0.25\textwidth]{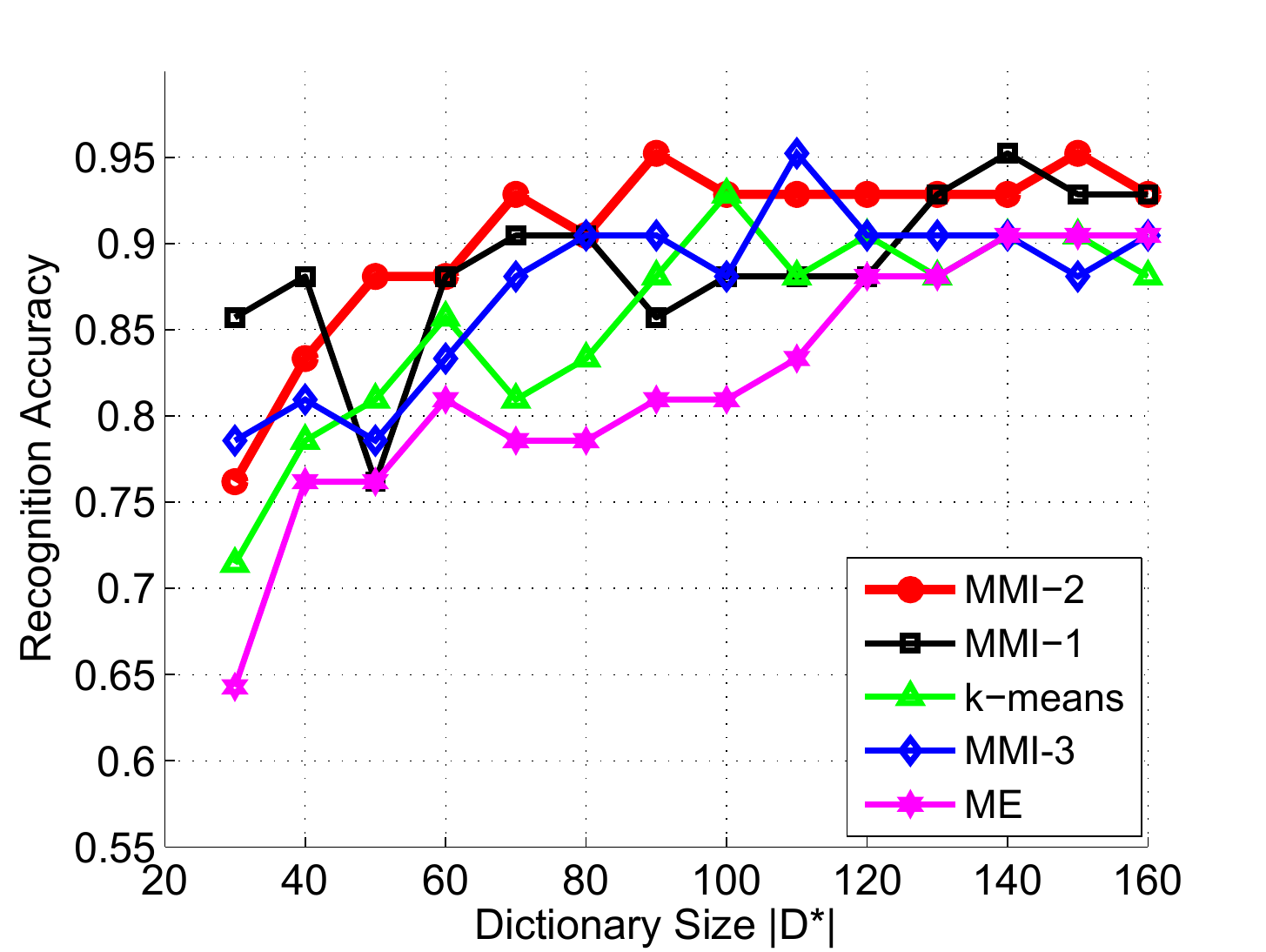}}
\caption{Recognition accuracy on the Keck gesture dataset with different features and dictionary sizes  (\emph{shape} and \emph{motion} are global features. \emph{STIP}~\cite{Laptev08} is a local feature.). The recognition accuracy using initial dictionary $D^o$: (a) 0.23 (b) 0.42 (c) 0.71 (d) 0.81. In all cases, the proposed MMI-2 (red line) outperforms the rest.
}
\label{Fig:gestureacc}
\end{figure*}

\begin{figure*} [ht]
\centering
\includegraphics[angle=0, height=0.09\textwidth, width=1\textwidth]{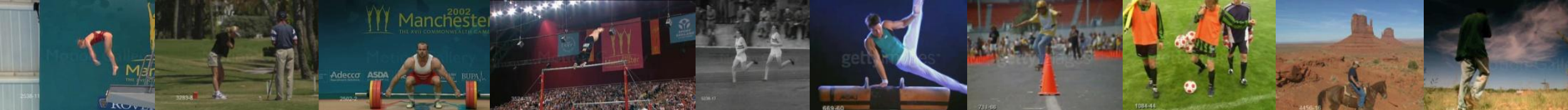}
\caption{Sample frames from the UCF sports action dataset. The actions include: diving, golfing, kicking, weight-lifting, horse-riding, running, skateboarding, swinging-1 (on the pommel horse and on the floor), swinging-2 (at the high bar), walking.}
\label{fig:ucf}
\end{figure*}

\begin{figure*} [ht]
\centering
\includegraphics[angle=0, height=0.45\textwidth, width=0.85\textwidth]{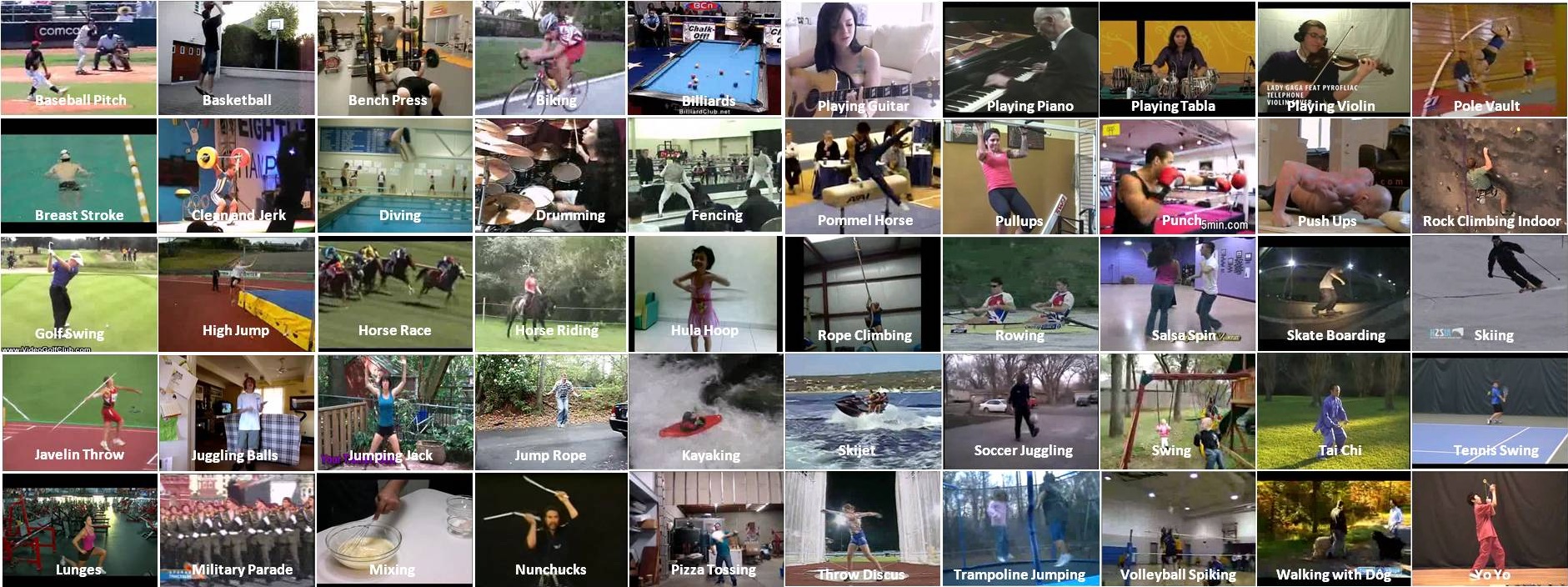}
\caption{Sample frames from the UCF50 action dataset. UCF50 is an action recognition dataset with 50 action categories, consisting of 6617 realistic videos taken from youtube.}
\label{Fig:ucf50}
\end{figure*}

\subsubsection{Dictionary Purity and Compactness}

Through K-SVD, we start with an initial 500 size dictionary using the shape feature (sparsity 30 is used). We then learned a 40 size dictionary $D^*$ from $D^o$ using 5 different approaches.
We let $\lambda=1$ in (\ref{eqt:maxmutualinfo2-greedy}) throughout the experiment.
To evaluate the discriminability and compactness of these learned dictionaries, we evaluate the \emph{purity} and \emph{compactness} measures as shown in Fig.~\ref{Fig:purity-compactness}. The purity is assessed by the histograms of the maximum probability observing a class given a dictionary atom, i.e., $max (P(L|d_i))$, and the compactness is assessed by the histograms of $D^{*T}D^*$. As each dictionary atom is $L_2$-normalized, $d_i^td_j \in [0,1]$ and
indicates the similarity between dictionary atoms $d_i$ and $d_j$.
Fig.~\ref{fig:purity} shows MMI-2 is most ``pure", as around $25\%$ of dictionary atoms learned by MMI-2 have 0.6-above probability to only associate with one of the classes. MMI-3 shows comparable purity to MMI-2 as the MI loss criteria used in MMI-3 does retain the class information during dictionary learning. However, as shown in Fig.~\ref{fig:compactness}, MMI-2 dictionary is much more compact, as only about $20\%$ MMI-2 dictionary atoms have $0.80$-above similarity. As expected, comparing to MMI-2, MMI-1 shows better compactness but much less purity.

 \subsubsection{Describing Unknown Actions}

We illustrate here how unknown actions can be described through a learned attribute dictionary.
We first obtain a 500 size initial shape dictionary $D^o$ using 11 out of 14 gesture classes, and keep \emph{flap}, \emph{stop both} and \emph{attention both} as unknown actions. We would expect a near perfect description to these unknown actions, as we notice these three classes are composed by attributes observed in the rest classes. For example, \emph{flap} is a two-arm gesture ``unseen" by the attribute dictionary, but its left-arm pattern is similar to \emph{turn left}, and right-arm is similar to \emph{turn right}.

As shown in Fig.~\ref{Fig:dic40},  we learned 40 size dictionaries using MMI-2, ME and MMI-3 respectively from $D^o$. Through visual observation, ME dictionary (Fig.~\ref{fig:dicme}) is most compact as dictionary atoms look less similar to each other. However, different from MMI-2 dictionary (Fig.~\ref{fig:dicno2}), it contains shapes mostly associated with the action start and end as discussed in Sec.~\ref{sec:learning}, which often results in high reconstruction errors shown in Fig.~\ref{fig:compositionplot}.  MMI-3 dictionary (Fig.~\ref{fig:dicmm}) only concerns about the discriminability, thus obvious redundancy can be observed in its dictionary. We can see from Fig.~\ref{fig:compositionplot}, though the action \emph{flap} is unknown to the dictionary, we still obtain a nearly perfect reconstruction through MMI-2, i.e.,  we can perfectly describe it using attributes in dictionary with corresponding sparse coefficients.

\subsubsection{Recognition Accuracy}
\label{sec:gestureacc}

In all of our experiments, we use the following classification schemes: when the global features, i.e., \emph{shape} and \emph{motion}, are used for attribute dictionaries,  we first adopt dynamic time warping (DTW) to align and measure the distance between two action sequences in the sparse code domain; then a $k$-NN classifier is used for recognition.
When the local feature \emph{STIP} \cite{Laptev08} is used,  DTW becomes not applicable, and we simply perform recognition using a $k$-NN classifier based on the sparse code histogram of each action sequence.

In Fig.~\ref{Fig:gestureacc}, we present the recognition accuracy on the Keck gesture dataset with different dictionaries sizes and over different global and local features. We use a leave-one-person-out setup, i.e., sequences performed by a person are left out, and report the average accuracy.  We choose an initial dictionary size $|D^o|$ to be twice the dimension of an input signal and sparsity 10 is used in this set of experiments.
In all cases, the proposed MMI-2 outperforms the rest. The sparse code noise has more effects on the DTW methods than the histogram method, thus, MMI-2 brings more improvements on global features over local features. The peak recognition accuracy obtained from MMI-2 is comparable to $92.86\%$ (motion),  $92.86\%$ (shape),  $95.24\%$ (shape and motion) reported in \cite{Lin09}.

As discussed, the near-optimality of our approach can be guaranteed if the initial dictionary size $|D^o|$ is sufficiently larger than $2|D^*|$.  We usually choose a size for $D^*$ to keep $|D^o|$ be $10$ to $20$ times larger. As shown in Fig.~\ref{Fig:gestureacc}, such dictionary size range usually produces good recognition performance. We can also decide $|D^*|$  when the MI increase in (\ref{eqt:maxmutualinfo2-greedy}) is below a predefined threshold, which can be obtained via cross validation from training data.

\subsection{Discriminability of Learned Action Attributes}

In this section, we further evaluate the discriminative power of learned action attributes using MMI-2.

\subsubsection{Recognizing Unknown Actions}

The Weizmann human action dataset contains 10 different actions: bend, jack, jump, pjump, run, side, skip, walk, wave1, wave2. Each action is performed by 9 different people.
We use the shape and the motion features for attribute dictionaries.
In the experiments on the Weizmann dataset,  we learn a $50$ size dictionary from a $1000$ size initial dictionary and the sparsity 10 is used.
When we use a leave-one-person-out setup, we obtain $100\%$ recognition accuracy for the Weizmann dataset.

To evaluate the recognition performance of attribute representation for unknown actions, we use a leave-one-action-out setup for dictionary learning, and then use a leave-one-person-out setup for recognition. In this way, one action class is kept unknown to the learned attribute dictionary, and its sparse representation using attributes learned from the rest classes is used for recognition. The recognition accuracy is shown in Table~\ref{tab:weizmannacc}.

It is interesting to notice from the second row of Table~\ref{tab:weizmannacc} that
only \emph{jump} can not be perfectly described using attributes learned from the rest 9 actions, i.e., \emph{jump} is described by a set of attributes not completely provided by the rest actions. By examining the dataset, it is easy to notice \emph{jump} does exhibit unique shapes and motion patterns.

As we see from the third row of the table, omitting attributes of the \emph{wave2}, i.e., the \emph{wave-two-hands} action, brings down the overall accuracy most. Further investigation tells us, when the \emph{wave2} attributes are not present,  such accuracy loss is caused by $33\%$ \emph{pjump} being misclassified as \emph{jack}, which means the attributes contributed by \emph{wave2} are useful to distinguish \emph{pjump} from \emph{jack}.
This makes great sense as \emph{jack} is very similar to \emph{pjump} but \emph{jack} contains additional \emph{wave-two-hands} pattern.

\begin{table*}[ht]
\centering
{\small
	\begin{tabular}{|l|llllllllll|}
	\hline
Unknown Action & bend & jack & jump & pjump & run & side & skip & walk & wave1 & wave2 \\
	\hline
\hline
Action Accuracy & 1.00 & 1.00 & 0.89 & 1.00 & 1.00 & 1.00 & 1.00 & 1.00 & 1.00 & 1.00 \\
\hline
Overall Accuracy & 1.00 & 1.00 & 0.98 & 0.98 & 1.00 & 1.00 & 1.00 & 0.99 & 0.97 & 0.94 \\		
	\hline
	\end{tabular}
}	
	\caption{Recognition accuracy on the Weizmann dataset using a leave-one-action-out setup for dictionary learning. The second row is the recognition accuracy on the unknown action, and the third row is the overall average accuracy over all classes given the unknown action. The second row reflects the importance of attributes learned from the rest actions to represent the unknown action, and the third row reflects the importance of attributes from the unknown action to represent the rest actions. }
	\label{tab:weizmannacc}
\end{table*}

\subsubsection{Recognizing Realistic Actions}

\begin{figure}
\centering
\includegraphics[angle=0, height=.3\textwidth]{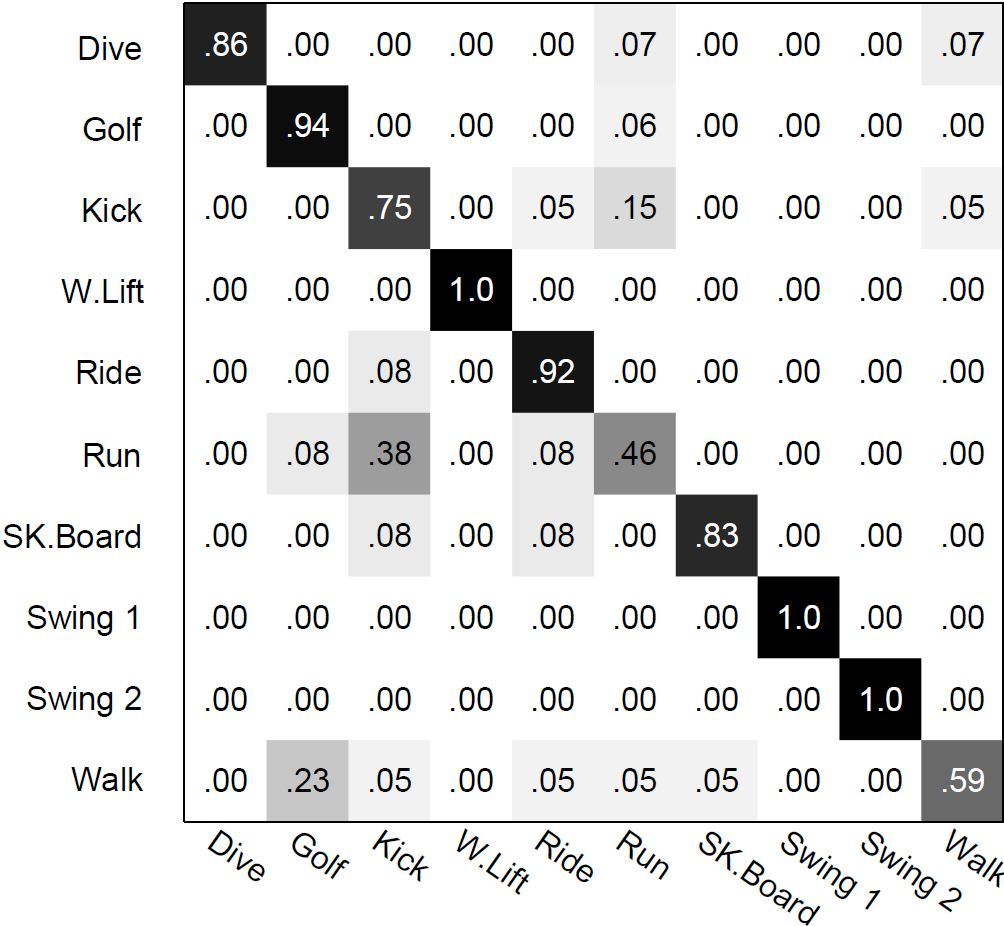}
\caption{Confusion matrix for UCF sports dataset}
\label{Fig:cm_ucf}
\end{figure}

The UCF sports dataset is a set of 150 broadcast sports videos and contains 10 different actions shown in Fig.~\ref{fig:ucf}. It is a challenging dataset with significant variations in scene content and viewpoints.
As the UCF dataset often involves multiple people in the scene, we use tracks from ground-truth annotations.
We use the HOG and the motion features for attribute dictionaries.
We learned a $60$ size dictionary from a $1200$ size initial dictionary and the sparsity 10 is used.
 We adopt a five-fold cross-validation setup.
With such basic features and a simple $k$-NN classifier, we obtain $83.6\%$ average recognition accuracy over the UCF sports action dataset, and the confusion matrix is shown in Fig.~\ref{Fig:cm_ucf}.

\subsection{Attribute dictionary on high-level features}

We learn our sparse attribute dictionary from features. As discussed in Sec.~\ref{sec:feature}, human actions are typically represented by low- or mid-level features, which contain little semantic meanings. Recent advances in action representations suggest the inclusion of semantic information for high-level action features. A promising high-level action feature, ActionBank, is introduced in \cite{actbank}.
The ActionBank representation is a concatenation of max-pooled detection features from many individual action detectors sampled broadly in a semantic space.
As reported in \cite{actbank}, the action recognition accuracy using ActionBank features is better than the state of the art,  better by 3.7\% on UCF Sports, and 10\% on UCF50.

In this section, we demonstrate that our learned action attributes can not only benefit from  but also enhance high-level features in terms of discriminability. We perform experiments on the UCF Sports and UCF50 action datasets.

We revisit the UCF sports dataset. Instead of the low-level HOG and motion features, we adopt the ActionBank high-level features for attribute dictionaries.
A $29930$ dimensional ActionBank feature is extracted for each action, and such feature is reduced to $128$ dimensions through PCA. Then, we learned a 40-sized attibute dictionary from a 128-sized initial dictionary and the sparsity 20 is used.
We use the same leave-one-out cross-validation setup as \cite{actbank} for action recognition. In order to emphasize the discriminability of learned action attributes, we adopt a simple $k$-NN classifier.

The recognition accuracies using high-level ActionBank features are reported in the second part of Table~\ref{tab:ucfsport-ab}. We obtain 90.7\% by using ActionBank features directly with a $k$-NN classifier. The recognition accuracy using the initial K-SVD dictionary on ActionBank features is 52.1\%. The recognition accuracy using the attribute dictionaries learned by MMI-1, MMI-2 and MMI-3 are 93.6\%, 91.5\% and 87.9\%. We made the following three observations: first, the proposed dictionary learning method significantly enhances dictionary discriminability (better by 41.5\% than the initial K-SVD dictionary). Second, the learned attributes using MMI-1 further improve the state of the art discriminability of ActionBank features (better by 3.0\%).
Third, discriminability improvements from considering class distribution during dictionary learning are less significant while using high-level features, comparing to low-level ones. This can be due to that high-level features like ActionBank have already encoded such semantic information, i.e., the feature appearance carries class information. Though MMI-2 significantly outperforms both MMI-2 and MMI-3 given low-level features, MMI-1 is preferred when high-level semantic features are used.

\begin{table}[h]
\centering
{\small
	\begin{tabular}{|l|l|}
	\hline
Method & Accuracy (\%) \\
	\hline
 \hline
Rodriguez et al. \cite{ucfsportdata} & 69.2 \\
Yeffet and Wolf \cite{yeffet-act} & 79.3 \\
MMI-2 (HOG\&motion) & 83.6 \\
Varma and Babu \cite{varma-act} & 85.2 \\
Wang et al. \cite{localst} &  85.6 \\
Le et al. \cite{le-act}  & 86.5 \\
Kovashka and Grauman \cite{kovashka-act}  &  87.3 \\
Wu et al. \cite{wu-act}  &  91.3 \\
\hline
\hline
K-SVD & 52.1 \\
MMI-3 & 87.9 \\
ActionBank & 90.7 \\
MMI-2  & 91.5 \\
MMI-1  & \textbf{93.7} \\
\hline
	\end{tabular}
}	
	\caption{Recognition accuracies on the UCF Sports dataset using high-level features.}
	\label{tab:ucfsport-ab}
\end{table}

We conduct another set of experiments using high-level features on the UCF50 action dataset. UCF50 is a very challenging action dataset with 50 action categories, consisting of 6617 realistic videos taken from youtube. Sample frames from the UCF50 action dataset are shown in Fig.~\ref{Fig:ucf50}.
A $14965$ dimensional ActionBank feature is first extracted for each action, and such feature is reduced to $512$ dimensions through PCA. Then, we learned a 128-sized dictionary from a 2048-sized initial dictionary and the sparsity 60 is used.
We use 5-fold group-wise cross-validation setup suggested in \cite{actbank} for action recognition. Again, we adopt a simple $k$-NN classifier.
We obtain 36.7\% by using ActionBank features directly with a $k$-NN classifier, and 41.5\% by using the MMI-1 attribute dictionaries learned from ActionBank features.
The learned action attributes further improve the discriminability of ActionBank features by 4.8\%.

\subsection{Action Sampling/Summarization using MMI-1}

This section presents experiments demonstrating action summarization using the proposed MMI-1 algorithm. We first use the MPEG shape dataset \cite{mpeg-shape} to provide an objective assessment of diversity and coverage enforced by the MMI-1 sampling scheme. Then we provide action summarization examples using the UCF sports dataset.

As discussed in Sec~\ref{sec:feature}, actions are described using features extracted from an action interest region. Global action features are typically shape-based or motion-based descriptors.
As video summarization often lacks of objective assessment schemes,
shape sampling provides an objective alternate to measure diversity and coverage of a sampling/summarization method.

\begin{figure} [h]
\centering
 \subfloat[10 classes from MPEG shape dataset] {\label{fig:mpeg_shape} \includegraphics[angle=0, height=0.5\textwidth, width=0.46\textwidth]{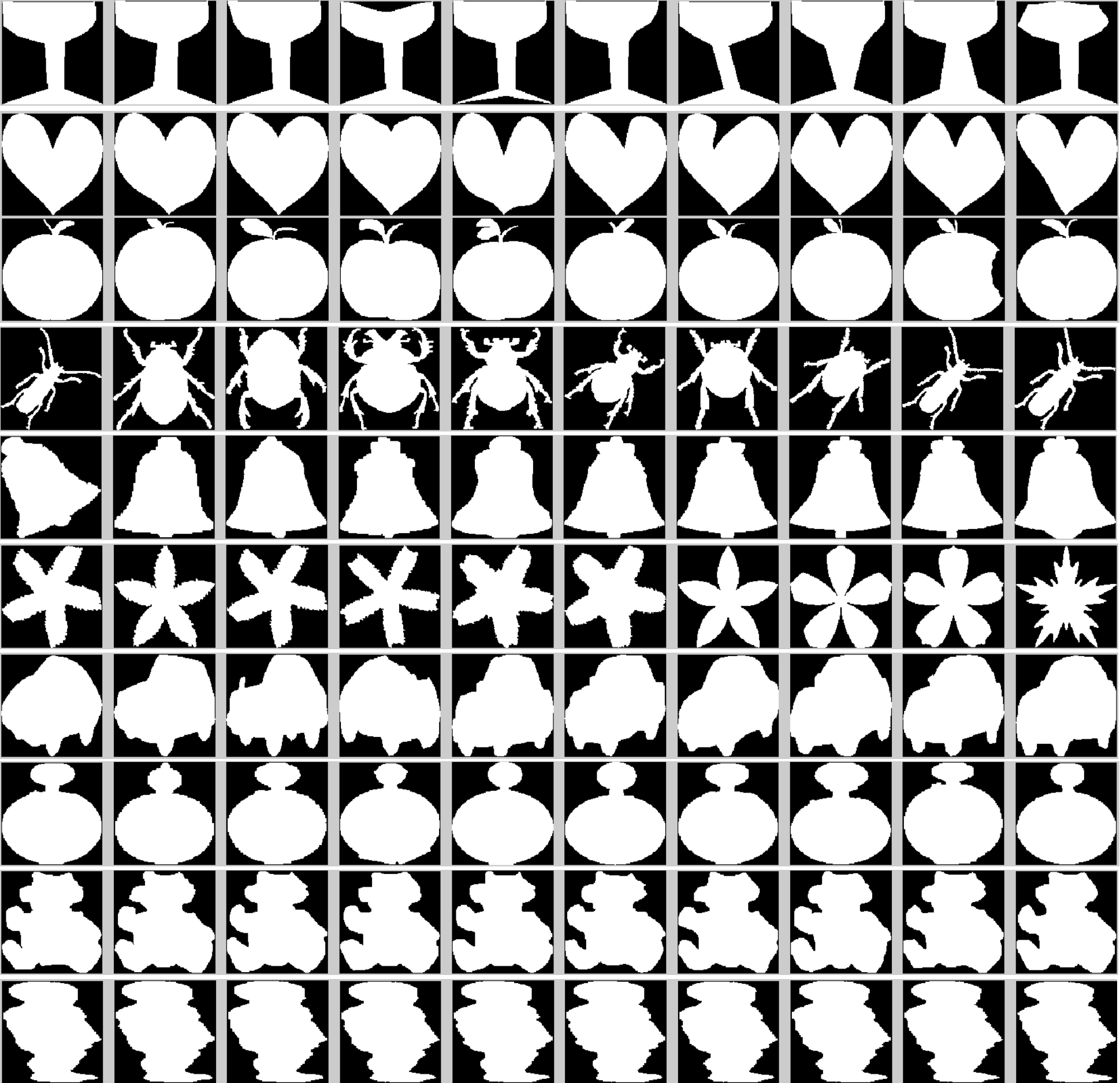} \hspace{13pt}} \\
  \subfloat[Top-10 shapes sampled using ME] {\label{fig:mpeg_me} \includegraphics[angle=0, height=0.05\textwidth, width=0.46\textwidth]{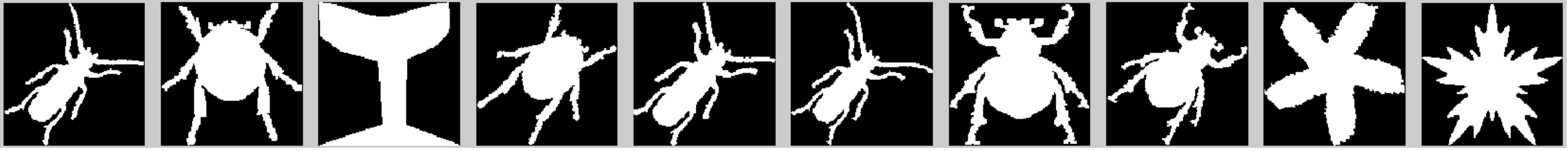}} \\
  \subfloat[Top-10 shapes sampled using k-means] {\label{fig:mpeg_kmean} \includegraphics[angle=0, height=0.05\textwidth, width=0.46\textwidth]{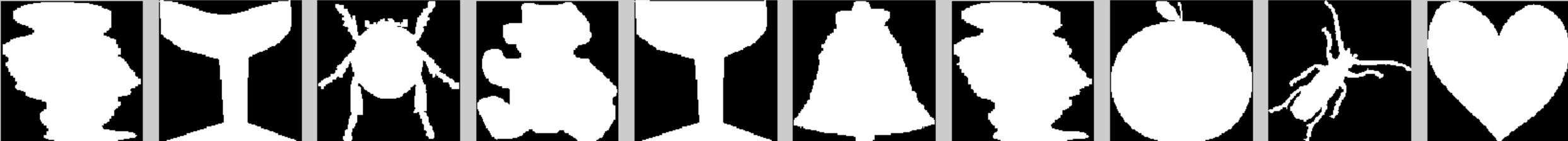}} \\
  \subfloat[Top-10 shapes sampled using the proposed MMI-1] {\label{fig:mpeg_mmi} \includegraphics[angle=0, height=0.05\textwidth, width=0.46\textwidth]{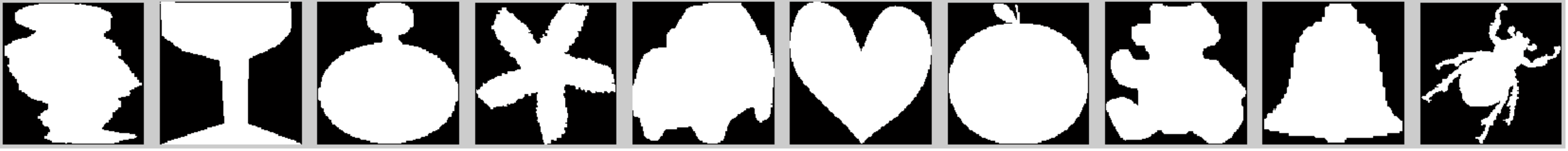}}
\caption{Shape sampling on the MPEG dataset.  The proposed MMI-1 method, which enforces both diversity and coverage criteria, retrieved all 10 shape classes.}
\label{Fig:mpeg7}
\end{figure}

We conducted shape sampling experiments on the MPEG dataset. This dataset contains 70 shape classes with 20 shapes each. As shown in Fig.~\ref{fig:mpeg_shape}, we use 10 classes with 10 shape each in our experiments.
To emphasize both diversity and coverage criteria, we keep our shape descriptor be variant to affine transformations. Thus, shapes with distinct rotation, scaling or translation are considered as outliers. The Top-10 shape sampling results using ME in Fig.~\ref{fig:mpeg_me}, which only considers diversity, retrieved 3 classes. The sampling results using k-means in Fig.~\ref{fig:mpeg_kmean}, which focuses on coverage, retrieved 7 classes.
As shown in Fig.~\ref{fig:mpeg_mmi}, the sampling results using the proposed MMI-1 method, which enforces both diversity and coverage criteria, retrieved all 10 classes.

\begin{figure*} [ht]
\centering
 \subfloat[A UCF sports sample dive sequence] {\label{fig:ucf-dive} \includegraphics[angle=0, height=0.8\textwidth, width=0.8\textwidth]{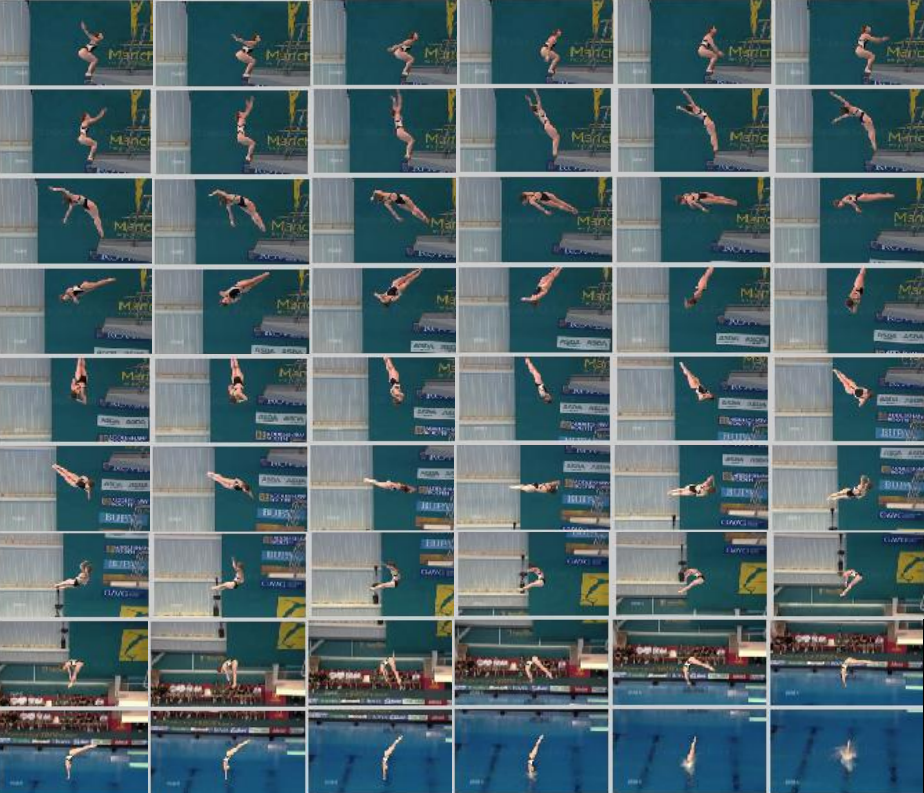} \hspace{13pt}}\\
  \subfloat[A dive action summary obtained using MMI-1 ] {\label{fig:dive-mmi} \includegraphics[angle=0, height=0.07\textwidth, width=0.8\textwidth]{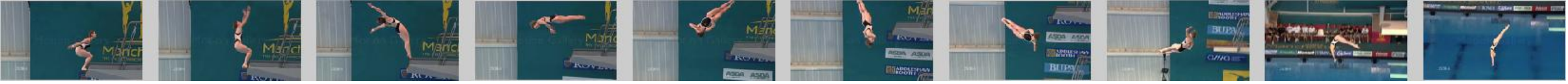} \hspace{13pt}}
\caption{An MMI-1 action summarization example using the UCF sports dataset}
\label{Fig:dive-summary}
\end{figure*}

In Fig.~\ref{Fig:dive-summary}, we provide an action summarization example using the proposed MMI-1 method.
For the dive sequence in Fig.~\ref{fig:ucf-dive}, we describe each frame of the action using both the HOG and the motion features. Then we sample Top-10 frames using MMI-1 and sort them by timestamps, as shown in Fig.~\ref{fig:dive-mmi}.  Through a subjective assessment, the dive action summarized using MMI-1 in Fig.~\ref{fig:dive-mmi} is compact yet representative.

\section{Conclusion}
We presented an attribute dictionary learning approach via information maximization for action recognition and summarization. By formulating the mutual information for appearance information and class distributions between the learned dictionary and the rest of dictionary space into an objective function, we can ensure the learned dictionary is both representative and discriminative. The objective function is optimized through a GP model proposed for sparse representation. The sparse representation for signals enable the use of kernels locality in GP to speed up the optimization process.  An action sequence is described through a set of action attributes, which enable both modeling and recognizing actions, even including ``unseen" human actions. Our future work includes how to automatically update the learned dictionary for a new action category.

{
\bibliographystyle{IEEEtran}
\bibliography{mainthesis}
}

\end{document}